\documentclass{article}

\usepackage{arxiv}

\usepackage[utf8]{inputenc} 
\usepackage[T1]{fontenc}    
\usepackage{hyperref}       
\usepackage{url}            
\usepackage{booktabs}       
\usepackage{amsfonts}       
\usepackage{nicefrac}       
\usepackage{microtype}      
\usepackage{lipsum}		
\usepackage{graphicx}
\usepackage{doi}
\usepackage{bbm}

\usepackage{amsmath,amssymb,amsfonts}
\usepackage{algorithmic}
\usepackage{graphicx}
\usepackage{textcomp}
\usepackage{biblatex}
\usepackage{algorithmic}
\def\BibTeX{{\rm B\kern-.05em{\sc i\kern-.025em b}\kern-.08em
    T\kern-.1667em\lower.7ex\hbox{E}\kern-.125emX}}
\usepackage{graphicx}
\usepackage{comment}
\usepackage{textcomp}
\usepackage{xcolor}
\usepackage{color}
\usepackage{mathtools}

\usepackage[inline]{enumitem}
\usepackage[ruled,vlined,linesnumbered]{algorithm2e}

\usepackage{multirow}
\usepackage{tabularx}
\usepackage{longtable}

\title{A  Survey on Optimal Transport for Machine Learning: Theory and Applications}

\author{ {\hspace{1mm}Luis Caicedo ~Torres}\\
	Department of Mathematics and Statistics\\
	Florida International University\\
	Miami, FL, 33199 \\
	\texttt{lcaic005@fiu.edu} \\
	\And
	{\hspace{1mm}Luiz Manella ~Pereira} \\
	Knight Foundation School of Computing and Information Sciences\\
	Florida International University, solid lab \\
	Miami, FL,33199 \\
	\texttt{lpere339@fiu.edu} \\
	\And
	{\hspace{1mm}M. Hadi ~Amini} \\
	Knight Foundation School of Computing and Information Sciences\\Sustainability, Optimization, and Learning for InterDependent networks laboratory (solid lab) \\
	Florida International University\\
	Miami, FL, 33199 \\
	\texttt{moamini@fiu.edu} \\
}

\date{}

\renewcommand{\shorttitle}{\textit{arXiv} Template}

\begin{document}
\maketitle

\begin{abstract}
	Optimal Transport (OT) theory has seen an increasing amount of attention from the computer science community due to its potency and relevance in modeling and machine learning. It introduces means that serve as powerful ways to compare probability distributions with each other, as well as producing optimal mappings to minimize cost functions. Therefor, it has been deployed in computer vision, improving image retrieval, image interpolation, and semantic correspondence algorithms, as well as other fields such as domain adaptation, natural language processing, and variational inference. In this  survey, we propose to convey the emerging promises of the optimal transport methods across various fields, as well as  future directions of study for OT in machine learning. We will begin by looking at the history of optimal transport and introducing the founders of this field. We then give a brief glance into the algorithms related to OT. Then, we will follow up with a mathematical formulation and the prerequisites to understand OT, these include Kantorovich duality, entropic regularization, KL Divergence, and Wassertein barycenters. Since OT is a computationally expensive problem, we then introduce the entropy-regularized version of computing optimal mappings, which allowed OT problems to become applicable in a wide range of machine learning problems. In fact, the methods generated from OT theory are competitive with the current state-of-the-art methods. The last portion of this survey will analyze papers that focus on the application of OT within the context of machine learning. We first cover computer vision problems; these include GANs, semantic correspondence, and convolutional Wasserstein distances. Furthermore, we follow this up by breaking down research papers that focus on graph learning, neural architecture search, document representation, and domain adaptation. We close the paper with a small section on future research. Of the recommendations presented, three main problems are fundamental to allow OT to become widely applicable but rely strongly on its mathematical formulation and thus are hardest to answer. Since OT is a novel method, there is plenty of space for new research, and with more and more competitive methods (either on an accuracy level or computational speed level) being created, the future of applied optimal transport is bright as it has become pervasive in machine learning.
\end{abstract}


\section{Introduction}

The Optimal Transport problem sits at the intersection of various fields, including probability theory, PDEs, geometry, and optimization theory. It has seen a natural progression in its theory from when Monge first posed the problem in 1781 \cite{monge1781memoire}. Now, it serves as a powerful tool due to its natural formulation in various contexts. It has recently seen a wide range of applications in computer science--most notably in computer vision, but also in natural language processing and other areas. Different elements such as the Convolutional Wasserstein Distance \cite{solomon2015convolutional} and the Minibatch Energy Distance \cite{arjovsky2017wasserstein} have made significant improvements on image interpolation, heat maps, and GANs. These are examples of some problems in machine learning that are being recast using Optimal Transport elements, such as Wasserstein distance being used as an error measure for comparing different probability distributions. We note the effectiveness with which optimal transport deals with both discrete and continuous problems and the easy transition between the two classes of problems. The powerful tools from convex geometry and optimization theory have made optimal transport more viable in applications. To that extent, we note the remarkable implementation of Sinkhorn's algorithm to significantly speed up computation of Wasserstein distances \cite{cuturi2013sinkhorn}.

Although the theory is well-developed \cite{villani2003topics}, much work is being made in determining the state-of-the-art algorithms for computing optimal transport plans under various conditions. In this survey, we explore the main tools from the theory and summarize some of the major advancements in its application. While it is not all-encompassing, we aim to provide an application-focused summary.

The rest of this paper is organized as follows: Section 2 provides an overview of algorithms from different applications and major breakthroughs in computation. Section 3 presents a brief history of the topic. Section 4 details some mathematical formalism. Section 5 reviews ways to overcome the computational challenges. Section 6 and on then explores applications of OT to different fields, most notably in GANs and general image processing. We then conclude with remarks and proposed directions and close with open problems. 

The interested reader can dive deeper into the rich OT material using some superb books such as \cite{villani2003topics}, \cite{villani2008optimal}, \cite{peyre2019computational}, \cite{santambrogio2015optimal}.

\section{OT Algorithms at a Glance}

\begin{table} [h]
\caption{OT Algorithms in Machine Learning Presented}
\label{table}
\setlength{\tabcolsep}{10pt}
\begin{tabular}{|p{85pt}|p{185pt}|p{90pt}|p{20pt}|}
\hline \rule{0pt}{2ex} 
Application& 
Publication& 
Metric Employed&
Year\\
\hline \rule{0pt}{2ex}    
Computations & 
Sinkhorn Entropy-Reg OT \cite{cuturi2013sinkhorn} & 
Ent-Reg W-Distance &
2013\\ 
\rule{0pt}{2ex} Computations & 
2-W Barycenters \cite{cuturi2014fast} & 
Ent-Reg W-Distance &
2014\\
\rule{0pt}{2ex} Comp. Vision & 
Conv-W Dist \cite{solomon2015convolutional} &  
Conv-W &
2015\\ 
\rule{0pt}{2ex} Comp. Vision & 
WGANs \cite{arjovsky2017wasserstein} & 
EMD &
2017\\
\rule{0pt}{2ex} Comp. Vision & 
OT-GAN \cite{salimans2018improving} &  
MED  &
2018\\
\rule{0pt}{2ex} Graphs & 
GWL \cite{kandasamy2018neural} &  
Gromov - W Dist &
2018\\
\rule{0pt}{2ex} Domain Adaptation & 
GCG \cite{flamary2016optimal} & 
Ent-Reg W-Distance &
2016 \\ \hline
\multicolumn{4}{p{444pt}}{An Overview of the algorithms presented in detail. Abbreviations used: Entropy Regularized Wasserstein Distance (Ent-Reg W-Distance), Minibatch Energy Distance (MED), Convolutional Wasserstein Distance (Conv-W), Gromov Wasserstein Distance (Gromov-W Dist) Earth Mover Distance (EMD), Domain Adaptation (Dom. Adap.), 2-Wasserstein (2-W), Gromov-Wasserstein Learning (GWL), Generalized Conditional Gradient (GCG) }
\end{tabular}
\label{tab1}
\end{table}

\section{History}
The central idea of Optimal Transport (OT) can be found in the work by French geometer Gaspard Monge. In his paper, \textit{Mémoire sur la théorie des déblais et
des remblais}, published in 1781, Monge asked the question: How do I move a pile of earth (some natural resource) to a target location with the least amount of effort, or cost \cite{monge1781memoire}? The idea was to find a better way of optimizing such cost that was not simply iterating through every possible permutation of supplier vs. receiver and choosing the one with the lowest cost. One of the major breakthroughs following Monge's work was by Russian mathematician Leonid Vitaliyevich Kantorovich who was the founder of linear programming. His research in optimal resource allocation, which earned him his Nobel Prize in Economics,  led him to study optimal coupling and duality, thereby recasting some parts of the OT problem into a linear programming problem. Kantorovich's work led to the renaming of optimal coupling between two probability measures as the Monge-Kantorovich problem.

After Kantorovich, the field of OT gained traction and its applications expanded to several fields. For example, while John Mather worked on  Lagrangian dynamical systems, he developed the theory of action-minimizing stationary measures in phase space, which led to the solution of certain Monge-Kantorovich problems \cite{mather1989minimal}. Although he did not make the connection between his work and OT, Buffoni and Bernard in their paper \textit{Optimal mass transportation and Mather theory} showed the existence of an optimal transport map while studying the "Monge transportation problem when the cost is the action associated to a Lagrangian function on a compact manifold \cite{bernard2004optimal}."

Several other names helped expand the field OT. For example, Yann Brenier introduced optimal coupling to his research in incompressible fluid mechanics, thus linking the two fields. Mike Cullen introduced OT in meteorology while working on semi-geostrophic equations. Both Brenier's and Cullen's work brought forth the notion that there is a connection, previously not expected, between OT and PDEs. Fields medalist Cédric Villani also contributed much to the field in connection with his work in statistical mechanics and the Boltzmann equation.

Recently, OT is being applied in several fields, including Machine Learning (ML). It started with image processing by utilizing color histograms of images (or gray images) and Wasserstein's distance to compute the similarity between images. Then, it was followed by shape recognition \cite{peleg1989unified, gangbo2000shape, ahmad2003geometry}. For example, in \textit{A Metric for Distributions with Applications to Image Databases}, Rubner et al. introduced a new distance between two distributions, called Earth Mover's Distance (EMD), which reflects the minimal amount of work that must be performed to transform one distribution into the other by moving "distribution mass" around \cite{rubner1998metric,  rubner2000earth}. Next, Haker et al. introduced a method for computing elastic registration and warping maps based on the Monge-Kantorovich theory of OT \cite{haker2003monge, haker2004optimal}.

Due to the important role of matrix factorization in ML, it was a natural progression to use OT as the divergence component of Nonnegative Matrix Factorization (NMF) \cite{sandler2011nonnegative}. In 2014, Solomon et al., looked at the applications of OT in semi supervised learning in their paper \textit{Wasserstein Propagation for Semi-Supervised Learning} \cite{solomon2014wasserstein}. Other applications have been utilizing OT in mappings between distributions; more specifically, a recent paper was published on using Wasserstein's metric in variational inference, which lies at the heart of ML \cite{ambrogioni2018wasserstein}.

More recently, researchers have made advancements in the theory of OT with Marco Cuturi proposing methods to solve approximations of the OT problems by introducing a regularization term \cite{cuturi2013sinkhorn}. The field is now more active than ever, with researchers extending the theories that work for low-dimensional ML problems into high-dimensional problems, bringing forth several complex theoretical and algorithmic questions \cite{santambrogiooptimal}.

\section{Mathematical Formalism}
\subsection{Problem Statement}

Given a connected compact Riemannian manifold $M$, \textit{Optimal Transport Plans} (OT plans) offer a way to mathematically formulate the mapping of one probability measure $\mu_0$ \emph{onto} another probability measure $\mu_1$. These plans $\pi$ are couplings that obey mass conservation laws and therefore belong to the set 

$$ \Pi(\mu_0 , \mu_1 ) = \{ \pi \in \text{Prob}(M \times M) | \pi(\cdot , M) = \mu_0 , \pi(M , \cdot) = \mu_1 \} $$

Here, $\Pi$ is meant to be the set of all joint probabilities that exhibit $\mu_0$ and $\mu_1$ as marginal distributions. The OT plan $\pi(x,y)$ seeks to transport mass from point $x$ to point $y$. This formulation allows for \emph{mass-splitting} which is to say that the optimal transport map can take portions of the mass at point $x$ to multiple points $y_i$. Kantorovich sought to rephrase the Monge question into a minimization of a linear functional 
\begin{equation}
    \pi \rightarrow \text{inf} \int_{M \times M} c(x,y) d \pi(x,y) 
\end{equation}
on the nonempty and convex $\Pi$ and appropriate cost function $c$. We note that some formulations accommodating multiple cost have also been proposed, e.g. \cite{scetbon2020handling}. Alternatively, these OT plans will minimize the distance between two measures denoted formally as the \emph{2-Wasserstein Distance}, where $d$ is a metric:

\begin{equation} \label{w2metric}
    W^{d}_{2} (\mu_0, \mu_1) =  \inf_{\pi \in \prod(\mu_0 , \mu_1)} \bigg( \int_{M \times M} d(x,y)^2 d \pi (x,y) \bigg)^{1/2}
\end{equation} 

This distance defines a metric\footnote{Here, we mean a metric in the mathematics sense, i.e. a function $d(\cdot,\cdot): M \times M \to \mathbb{R}_+$ that is positive definite, symmetric, and subadditive on a metrizable space $M$. See Appendix A for more details.} as shown in Villani's book \cite{villani2003topics}. This distance metric will be integral to applications as we will see that it offers a new way to define loss functions. The goal is to find, or approximate, the optimal transport plan, $\pi$. 

\subsection{Kantorovich Duality}

Duality arguments are central to both the theoretical and numerical arguments in the OT framework. Kantorovich noticed that the minimization of the linear functional problem emits a dual problem. Here, let $c$ denote a lower semicontinuous cost function, $\mu_0$ and $\mu_1$ denote marginal probabilities, and $\Pi$ be the set of all probability measures on $M \times M$ which emit $\mu_0$ and $\mu_1$ as marginals. Then, for continuous $\phi(x), \psi(y)$, we have that 
\begin{equation}
    \inf\limits_{\Pi(\mu_0,\mu_1)} \int_{M \times M} c(x,y) d \pi(x,y) = \sup\limits_{\phi, \psi} \int_M \phi(x) d\mu_0 + \int_M \psi(y) d\mu_1
\end{equation}
The right-hand side of the equation is known as the \emph{dual problem} of the minimization problem and is a very useful tool in proving consequences regarding optimal transport maps. A proof of this result, along with further discussion, can be found in \cite{villani2003topics}.

\subsection{Entropic Regularization}

We can define the entropy of a coupling on $M \times M$ by the negative energy functional coming from information theory: 

\begin{equation}\label{entropy}
    H(\pi) = - \int \int_{M \times M} \pi (x,y) \ln(\pi(x,y)) dx dy
\end{equation}

This entropy essentially tracks information loss of a given estimate versus the true value as it proves a lower bound for the square loss error. Then, we can consider the entropy-regularized Wasserstein distance:

\begin{equation} \label{w2regmetric}
    W^2_{2,\gamma} (\mu_0 , \mu_1 ) =  \inf_{\pi \in \Pi(\mu_0 , \mu_1)} \bigg[ \int \int_{M \times M} d(x,y)^2 d\pi (x,y) - \gamma H(\pi) \bigg]
\end{equation} 

Cuturi proved this regularized distance offers a transport plan that is more spread out and also offers much faster computational convergence convergence \cite{cuturi2013sinkhorn}. This computational breakthrough will be pivotal in the tractability of Wasserstein-distance dependent algorithms. 

\subsection{KL Divergence} 

A lot of results for optimal transport maps can be related to the familiar KL divergence. If we define $p(x)$ and $q(x)$ as probability distributions given a random variable x over a manifold of distributions, then we define the KL divergence as:
\begin{equation} 
\label{kldiv} D_{KL}(p(x)|q(x)) \coloneqq \int p(x) \bigg(\ln \frac{p(x)}{q(x)}\bigg) dx 
\end{equation}

\subsection{Wasserstein barycenters} 

The barycenter problem is central to the interpolation of points in Euclidean space. Agueh and Carlier present the analog in Wasserstein space, proving its existence, uniqueness, and providing characterizations \cite{agueh2011barycenters}. The analog is presented as the solution to the minimzation of a convex combination problem 
\begin{equation} \label{wassbary}
    \inf_{\mu} \sum\limits_{i=1}^{p} \alpha_i W^2_2(\mu_i,\mu) 
\end{equation}
where $\mu_i$ are probability measures and the $\alpha_i$'s, known as barycentric coordinates, are nonnegative and sum to unity. These conclusions are derived from considering the problem dual to the problem and desirable properties of the Lengendre-Fenchel transform as well as conclusions from convex geometry. These barycenters are also uniquely characterized in relation to Brennier maps which offers direct formulation as a push forward operator. Barycenteres will play a major role in applications such as the interpolation of images under transport maps as in \cite{solomon2015convolutional}. Computing these barycenters is discussed in the Computational Challenges section. 

\section{Computational Challenges}

One the biggest challenges in the implementation of optimal transport has been its computational cost. One widely used implementation of Sinkhorn's algorithm was formulated by Cuturi significantly decreased computation cost \cite{cuturi2013sinkhorn}. In the following, $KL$ denotes the Kullback-Leibler divergence, $U$ denotes the transport polytope of transport plans $P$ that emit $r$ and $c$ as marginal distributions: $U(r,c) = \{ P \in \mathbb{R}^{d \times d}_+ | P \mathbbm{1}_d = r, P^T \mathbbm{1}_d = c \}$; and $U_\alpha(r,c) = \{  P\in U(r,c) | KL(P | rc^T) \leq \alpha\} $. We present the discrete version as opposed to the continuous analog presented in equation (\ref{w2regmetric}). Define the Sinkhorn distance as 
\begin{equation} \label{sinkhorn}
d_{M,\alpha} \coloneqq \min\limits_{P \in U_\alpha(r,c)} \langle P,M \rangle \end{equation}
Then we can introduce an entropy regularization argument stated in a Lagrangian for $\lambda >0 $: 

\begin{equation} \begin{aligned} \label{Regent} d_M^\lambda (r,c) \coloneqq & \langle P^\lambda , M \rangle,  \quad \\ \text{where} \quad P^\lambda = \text{argmin}_{P \in U(r,c)} &\langle P,M \rangle - \frac{1}{\lambda} h(P) \end{aligned}
\end{equation} 

where $h(P) = - \sum\limits_{i,j=1}^d p_{ij} \log(p_{ij})$ is the entropy of $P$. Then, Sinkhorn's famed algorithm for finding the the minimum, which we know from the general theory will be found on one of the vertices of the polytope, will serve as a proper approximation tool as seen in Algorithm \ref{alg:RegEnt}. Here, a main result proved by Cuturi is used which states that the solution $P^\lambda$ is unique and, moreover, has the particular form of $P^\lambda = \textit{diag}(u) K \textit{diag}(v)$, where $u,v$ are two nonnegative vectors that are unique up to constants and $K = e^{- \lambda M}$ denotes the matrix exponential of $ - \lambda M$. This result is pivotal in further speeding up the computation of (\ref{Regent}). This type of result is also commonly used as in, for example, \cite{solomon2014wasserstein} which is explored in (\ref{cwd}).
\LinesNumberedHidden{
\begin{algorithm} 
\DontPrintSemicolon
\caption{\textbf{Computation of Entropy-Regularized} \\ $d = [d^\lambda_M(r, c_1) ,d^\lambda_M(r, c_2), ... , d^\lambda_M(r, c_N)]$ }
\label{alg:RegEnt}

Input: $M,\lambda, r, C = [c_1, ... , c_N]$ \\ $I = (r>0) ; r = r(I) ; M = M(I,:); K = exp(- \lambda M)$ \\ $u = ones(length(r),N) / length(r)$ \\ $\hat{K} = diag(1./r) K$ \\
While $u$ changes or any stopping criterion Do \\
$\quad u = 1./(\hat{K}(C./(K^Tu)))$ \\
end while \\
$v = C./ (K^T u)$ \\
$d = sum(u.*((K.*M)v))$
\end{algorithm}}

The implementation of Sinkhorn's algorithm to find optimal transport maps has improved the general tractability OT algorithms. We note the improvement on the problem of computing barycenters in Wasserstein space made by Cuturi and Doucet in \cite{cuturi2014fast} where they prove the polyhedral convexity of a function that is like a discrete version of (\ref{wassbary}) 
$$ f(r,X) = \frac{1}{N} \sum\limits_{i=1}^{N} d(r,c_i,M_{XY_i}) $$ where $d(\cdot,\cdot)$ is the previously defined Sinkhorn distance (\ref{sinkhorn}), and $r,c$ are the marginal probabilities. $M_{XY}$ is the pairwise distance matrix. Here, the problem of the optimal p is phrased using the dual linear programming from known as the dual optimal transport problem: 
$$ d(r,c,M) = \max_{(\alpha,\beta) \in C_M} \alpha^T r + \beta^T c$$ 
where $C_M$ is the polyhedron of dual variables 
$$C_M = \{ (\alpha,\beta) \in \mathbb{R}^{n+m} | \alpha_i + \beta_j \leq m_{ij}\}$$
This problem then has a solution and the computation of the barycenters centers around this. 

While the theoretical groundwork for optimal transport has been laid, efficient algorithms are still needed for it to be implemented in large scale. Genevay, \emph{et al.} formulate stochastic descent methods for large scale computations, making use of the duality arguments previously presented along with entropic regularization for various cases in \cite{genevay2016stochastic}. Then, Sinkhorn's algorithm will play an important role in the discrete case while the continuous case is very elegantly dealt with using reproducing kernel Hilbert spaces. 

For a complete discussion of the numerical methods associated with the OT problem as well as other relevant algorithms, see \cite{vialard2019elementary,merigot2020optimal}.

\section{Applications} 

Here, we hope to bring light to some of the many applications of OT within a machine learning setting. 
\subsection{Computer Vision}

OT finds a natural formulation within the context of computer vision. The common method is to make a probability measure out of color histograms relating to the image. Then, one can find a dissimilarity measure between the images using the Wasserstein distance. An early formulations of OT in computer visions can be in \cite{rubner2000earth} and those relating to the Earth Mover's Distance (EMD) which acts as a slighty different discrete version of the 1-Wasserstein distance. A formulation of the EMD on discrete surfaces can be found in \cite{solomon2014earth}. In the forthcoming, we note a use of OT in improving GANs and the Convolutional Wasserstein Distances which serve well for image interpolation. 

\subsubsection{OT Meets GANs} 

Multiple attempts have been made to improve GANs using optimal transport. Arjovski \emph{et al.} recast the GANs problem into an OT theory problem \cite{arjovsky2017wasserstein}. OT lends itself well to the GANs problem of learning models that generate data like images or text with a distribution that is similar to that of training data. Here in WGANs, we can take two probability measures $\mu_0,\mu_1 \in M$ with $\mu_1$ being the distribution of a locally Lipschitz $g_\theta(Z)$ acting as a neural network with \emph{nice} convergence properties and with $Z$ a random variable with density $\rho$ and $g_\theta$  and the Kantorovich-Rubinstein duality gives
$$ W(\mu_0,\mu_1) = \sup\limits_{||f|| \leq 1} \mathbb{E}_{x\sim \mu_0} [f(x)] - \mathbb{E}_{x\sim\mu_1} [f(x)] $$ 
with supremum taken over all Lipschitz continuous functions $f:M \to \mathbb{R}$. It is shown here that there is a solution to this problem with relevant gradient
$$ \nabla_\theta W(\mu_0,\mu_1) = - \mathbb{E}_{z \sim \rho} [\nabla_\theta f(g_\theta(z))]$$ 
wherever both are well-defined. This formulation poses an alternative to the classical GANs and it is found to be more stable, specially when dealing with lower dimensional data, than its counterparts. 

 We also note the progress made by Salimans \emph{et al.} \cite{salimans2018improving} where they improve upon the idea of mini-batches \cite{genevay2017learning} and using energy functionals \cite{BellemareDDMLHM17} to introduce an OT variant using the W-distance named the Minibatch Energy Distance: 
 \begin{align*} D^2_{MED} (\mu_0, \mu_1) = 2 \mathbb{E}[W_c(X,Y)] - \mathbb{E}[W_c(X,X')] - \mathbb{E}[W_c(Y,Y')] \end{align*}
 where $X,X'$ are sampled mini-batches from $\mu_0$ and $Y,Y'$ are sampled mini-batches from $\mu_1$ and $c$ is the optimal transport function that is learned adversarially through the alternating gradient descent common to GANs. These algorithms are seeing a greater statistical consistency.
 
\subsubsection{Semantic Correspondence} OT is one of the few, if not the only, method that deals with mass-splitting phenomenon which commonly occurs in establishing dense correspondence across semantically similar images. This occurrence is in the form of a many-to-one matching in the assignment of pixels from a source of pixels to a target pixel as well as a one-to-many matching of the same type. The one-to-one matching problem can be recast as an OT problem as done in \cite{liu2020semantic}. Liu \emph{et al.} replace it with maximizing a total correlation where the optimal matching probability is denoted as
$$ P^* = \text{argmax}_{P} \sum\limits_{i,j} P_{ij} C_{ij}$$
where $P \in \mathbb{R}^{n \times m}_+,  P \mathbbm{1}_n = r, P^T \mathbbm{1}_m = c$ and $r,c$ are marginals in the same vein as in the section on computational challenges. Then, we can call $M = 1-C$ to be the cost matrix. Then, the problem becomes the optimal transport problem 
$$ P^* = \text{argmin}_P \sum_{i,j} P{ij} M_{ij} $$
where $P \in \mathbb{R}^{n \times m}_+,  P \mathbbm{1}_n = r, P^T \mathbbm{1}_m = c$. This problem can then be solved using known algorithms, like those proposed in the computation challenges section. Using the percentage of correct keypoints (PCK) evaluation metric, their proposed algorithm outperformed state-of -the-art algorithms by 7.4 (or 26\%), making it a huge improvement over other methods.

\subsubsection{Convolutional Wasserstein Distances} \label{cwd}
In \cite{solomon2015convolutional}, Solomon \emph{et al.} propose an algorithm for approximating optimal transport distances across geometric domains. Here, they make use of the entropy-regularized Wasserstein distance given by (\ref{w2regmetric}) for its computational advantages discussed in the Computational Challenges section: 
\begin{equation} \label{w2regmetric2} \begin{aligned}
    W^2_{2,\gamma} (\mu_0 , \mu_1 ) =  \inf\limits_{\pi \in \Pi(\mu_0 , \mu_1)} \bigg[ \int_{M \times M} d(x,y)^2 d\pi (x,y)  - \gamma H(\pi) \bigg] \end{aligned}
\end{equation} 
They use Varadhan's formula \cite{varadhan1967behavior} to approximate the distance $d(x,y)$ by transferring heat from $x$ to $y$ over a short time interval: 
$$d(x,y)^2 = \lim\limits_{t \to 0} [-2t \ln H_t(x,y) ] $$ where $H_t$ is the heat kernel associated to the geodesic distance $d(x,y)$. Then, we can use this value in a kernel defined by $K_\gamma (x,y) = e^{- \frac{d(x,y)^2}{\gamma}}$. We can conclude through algebraic manipulations that 
$$ W_{2,\gamma}^2(\mu_0,\mu_1) = \gamma [1 + \min\limits_{\pi \in \Pi} KL(\pi | K_\gamma)] $$ where $KL$ denotes the K-L divergence (\ref{kldiv}). Then, in order to compute the convolutional distances, we can discretize the domain $M$ with function and density vectors $\mathbf{f} \in \mathbb{R}^n$. Then, define area weights vector $\mathbf{a} \in \mathbb{R}^n_+$ with $\mathbf{a}^T \mathbbm{1} = 1$ and a symmetric matrix $\mathbf{H}_t$ discretizing $H_t$ such that 
$$ \int_M f(x) dx \approx \mathbf{a}^T \mathbf{f} \quad \text{and} \quad \int_M f(y) H_t(\cdot,y)dy \approx \mathbf{H}_t(\mathbf{a} \otimes \mathbf{f} ) $$
Thus we are ready to compute the convolutional Wasserstein distance as in Algorithm \ref{alg:convwass2}.

\begin{algorithm} 

\DontPrintSemicolon
\caption{\textbf{Convolutional Wasserstein Distance}}
\label{alg:convwass2}

Input: $\mu_0 , \mu_1 , H_t , a, \gamma$ \;
Sinkhorn Iterations: \;
$\mathbf{v}, \mathbf{w} \leftarrow 1$ \;
for $i = 1,2,3,...$ \;
\quad $\mathbf{v} \leftarrow \mu_0 ./ \mathbf{H}_t(\mathbf{a}.*\mathbf{w})$ \;
\quad $\mathbf{w} \leftarrow \mu_1 ./ \mathbf{H}_t(\mathbf{a}.*\mathbf{v})$ \;
KL Divergence: \;
Return $\gamma \mathbf{a}^t[(\mu_0.* \ln(\mathbf{v})) + (\mu_1 .* \ln(\mathbf{w})]$

\end{algorithm}

We note the authors' use of the Convolution Wasserstein Distance along with barycenters in the Wasserstein space to implement an image interpolation algorithm.

\subsection{Graphs} \label{Graphs}
The OT problem also lends itself to the formulation of dissimilarity measures within different contexts. In \cite{kolouri2020wasserstein}, authors developed a fast framework, referred to as WEGL (Wasserstein Embedding for Graph Learning), to embed graphs in a vector space.

We find that analogs of the dissimilarity measures can be defined on graphs and manifolds where the source manifold and target manifolds need not be the same. 
In \cite{xu2019gromov}, Xu et al. propose a new method to solve the joint problem of learning embeddings for associated graph nodes and graph matching. This is done using a regularized Gromov-Wasserstein discrepancy when computing the levels of dissimilarity between graphs. The computed distance allows us to study the topology each of the spaces.
The Gromov-Wasserstein discrepancy was proposed by Peyre as a succession to the Gromov-Wasserstein distance which is defined as follows:
\\\\
\textbf{Definition:}
Let $(X,d_X,\mu_{X})$ and $(Y,d_Y,\mu_{Y})$ be two metric measure spaces, where $(X,d_X)$ is a compact metric space and $\mu_X$ is a probability measure on X (with $(Y,d_Y,\mu_{Y})$ defined in the same way). The Gromov Wasserstein distance $d_{GW}(\mu_X,\mu_Y)$ is defined as
\\
\begin{equation*}
\inf\limits_{\pi \in \Pi (\mu_X, \mu_Y)}\int\limits_{X \times Y} \int\limits_{X \times Y} L(x,y,x',y')d\pi(x,y)d\pi(x',y'),
\end{equation*}

where $L(x,y,x',y') = |d_X(x,x') - d_Y(y,y')|$ is the loss function and $\Pi (\mu_X,\mu_Y)$ is the set of all probability measures on X x Y with $\mu_X$ and $\mu_Y$ as marginals. We note that the loss function could be continuous depending on the topology the metric space $X$ is endowed with. At the very least, we would want it to be $\pi$-measurable. 
\\\\
When $d_x$ and $d_y$ are replaced with dissimilarity measurements rather than strict distance metrics and the loss function \emph{L} is defined more flexibly, the GW distance can be relaxed to the \emph{discrepancy}. From graph theory, a graph is represented by its vertices and edges, \emph{G(V,E)}. If we let a metric-measure space be defined by the pair \emph{\textbf{(C,$\mu$)}}, then we can define the Gromov-Wasserstein discrepancy between two spaces, \emph{\textbf{($C_s,\mu_s$)}} and \emph{\textbf{($C_t,\mu_t$)}}, as:
\begin{equation*} \begin{aligned}
    d_{GW}(\mu_s,\mu_t) &= \min_{T\in \pi (\mu_s, \mu_t)}\sum_{i,j,i',j'}L(c^{s}_{ij},c^{t}_{i'j'})Tii'Tjj'
    \\ &= \min_{T\in \pi (\mu_s, \mu_t)}\langle L(C_s,C_t,T),T \rangle \end{aligned}
\end{equation*}
In order to learn the mapping that includes the correspondence between graphs and also the node embeddings, Xu et al. proposed the regularized GW discrepancy:
\begin{equation*}\begin{aligned}
\min\limits_{X_s,X_t}\min\limits_{T\in\Pi(\mu_s,\mu_t)}\langle L(C_s(X_s),C_t(X_t),T),T\rangle +\alpha\langle K(X_s,X_t),T\rangle + \beta R (X_s,X_t)
\end{aligned}
\end{equation*}

To solve this problem, the authors present Algorithm \ref{alg:gwl}.

\begin{algorithm}
\DontPrintSemicolon
\caption{\textbf{Gromov-Wasserstein Learning (GWL)}}
\label{alg:gwl}

Input: $\{C_s,C_t\}$, $\{\mu_s$,$\mu_t\}$, $\beta$, $\gamma$, the dimension D, the number of outer/inner iterations $\{M,N\}$. \;
Output: $X_s$, $X_t$, and $\hat{T}$\;
Initialize $X_s^{(0)}$, $X_t^{(0)}$ randomly, $\hat{T}^{(0)}=\mu_s \mu_t^T$.\;
For $m=0 : M-1$:\; 
\quad Set $\alpha_m = \frac{m}{M}$.\;
\quad For $n=0 : N-1$\;
\quad \quad Update optimal transport $\hat{T}^{(m+1)}$\;
\quad Obtain $X_s^{(m+1)}$, $X_t^{(m+1)}$\;
$X_s = X_s^{(M)}, X_t = X_t^{(M)}$ and $\hat{T}=\hat{T}^{(M)}.$\;
Graph matching:\;
Initialize correspondence set $P=\emptyset$\;
For $v_i \in V_s$\;
\quad $j = \mathrm{arg max}_j \hat{T}_{ij}. P=P\bigcup\{ (v_i \in V_s,v_j \in V_t)\}$.
\end{algorithm}

The proposed methodology produced matching results that are better than all other comparable methods and opens the opportunity for the improvement of well-known systems (i.e. recommendation systems).

We note that the Gromov-Wasserstein discrepancy can also be used to improve GANs, as is done in \cite{bunne2019learning}. Here, Bunne, et al., adapt the generative model to use the Gromov-Wasserstein discrepancy to perform GANs across different types of data. 

\subsection{Neural Architecture Search}
In this section we will look at the following paper: \emph{Neural Architecture Search
with Bayesian Optimisation and Optimal Transport} \cite{kandasamy2018neural}.
Bayesian Optimization (BO) refers to a set of methods used for optimization of a function $f$, thus making it perfect for solving the \emph{model selection} problem over the space of neural architectures. The difficulty posed in BO when dealing with network architecture is figuring out how to quantify \emph{(dis)similarity} between any two networks. To do this, the authors developed what they call a (pseudo-)distance for neural network architectures, called OTMANN (Optimal Transport Metrics for Architectures of Neural Networks). Then, to perform BO over neural network architectures, they created NASBOT, or Neural Architecture Search with Bayesian Optimization and Optimal Transport. To understand their formulation, we first look at the following definitions and terms. First, a Gaussian process is a random process characterized by an expectation function (mean function)  $\mu: \chi \rightarrow \mathbb{R}$ and a covariance (kernel) $\kappa = \chi^2 \rightarrow \mathbb{R}$. In the context of architecture search, having a large $\kappa (x,x')$, where $x,x' \in \chi$ and $\kappa(x,x')$ is the measure of similarity so that $f(x)$ and $f(x')$ are highly correlated; implying the GP imposes a smoothness condition on $f:\chi \rightarrow \mathbb{R}$. Next, the authors view a neural network (NN) as a graph whose vertices are the layers of the network $G =(L,E)$, where $L$ is a set of layers and $E$ the directed edges. Edges are denoted by a pair of layers, $(u,v)\in E$. A layer $u\in L$ is equipped with a layer label $ll(u)$, which denotes the type of operations performed at layer $u$ (i.e. $ll(1) = conv3$ means 3x3 convolutions). Then, the attribute $lu$ denotes the number of computational units in a layer. Furthermore, each network has \emph{decision layers}, which are used to obtain the predictions of the network. When networks have more than one decision layer, one considers the average of the output given by each layer. Lastly, each network has an input and output layer, $u_{in}$ and $u_{op}$ respectively; any other layer is denoted as a \emph{processing layer}.
\\
Using the definitions above, the authors describe the distance for neural architectures as $d:\chi^2 \rightarrow \mathbb{R}_+$; with the goal of obtaining a kernel for the GP where $\kappa(x,x')=exp(-\beta d(x,x')^p)$, given that $\beta,p\in\mathbb{R}_+$. We first look at the OTMANN distance. OTMANN is defined as the minimum of a matching scheme which attempts to match the computation at the layers of one network to the layers of another, where penalties occur given that different types of operations appear in matched layers. The OTMANN distance is that which minimizes said penalties. Given two networks $G_1(L_1,E_1)$ and $G_2(L_2,E_2)$ with $n_1, n_2$ layers respectively, the OTMANN distance is computed by solving the following optimization problem:
\begin{align*}
        \underset{Z}{\text{minimize}} \hspace{8pt}  \phi_{lmm}(Z) + \phi_{nas}(Z) +\nu_{str}\phi_{str}(Z) \\
        \text{subject to} \sum\limits_{j\in L_2}Z_{ij}\leq lm(i), \sum\limits_{i\in L_1} Z_{ij} \leq lm(j), \forall i,j
\end{align*}

In the above equation, $\phi_{lmm}$ is the label mismatch penalty, $\phi_{str}$ is the structural term penalty, $\phi_{nas}$ is the non-assigment penalty, $Z \in \mathbb{R}^{n_1 x n_2}$ denoting hte maount of mass matched between layer $i\in G_1$ and $j\in G_2$, $l_m: L \rightarrow \mathbb{R}_+$ is a layer mass, and lastly $\nu_{str} > 0$ determines the trade-off between the structural term and other terms. This problem can be formulated as an Optimal Transport problem and is proved in the appendix of the paper.\\
Next, we look at NASBOT. The goal here is to use the kernel $\kappa$, as previously mentioned, to define the neural architectures and to find a method to optimize the acquisition function:
\begin{equation*} \begin{aligned}
    \phi_t(x) = \mathbb{E}[\max &\{ 0, f(x) -\tau_{t-1}\}|\{(x_i,y_i)\}_{i=1}^{t-1} ], \\ \tau_{t-1} &= \underset{i\leq t-1}{\text{argmax}}\hspace{4pt} f(x_i)
    \end{aligned}
\end{equation*}

The authors solve this optimization problem using an evolutionary algorithm, whose solution leads to the creation of NASBOT. Detailed explanations on the algorithm and the methodology onto which the optimization was solved can be found in the appendix of the original paper.
After running an experiment to compare NASBOT against known methods, the authors show that NASBOT consistently had the smallest cross validation mean squared error. For the interested reader, there are illustrations for the best architectures found for the problem posed in the experiment proposed. 

\subsection{Document Representation}
In this section we will look at the following paper: \emph{Hierarchical Optimal Transport
for Document Representation} \cite{yurochkin2019hierarchical}.
In this paper, Yurochkin, \emph{et al.} combine hierarchical latent structures from topic models with geometry from word embeddings. \emph{Hierarchical} optimal topic transport document distances, referred to as HOTT, this method combines language information (via word embeddings) with topic distributions from latent Dirichlet allocation (LDA) to measure the similarities between documents. Given documents $d^1$ and $d^2$, HOTT is defined as:
\begin{equation}
    HOTT(d^1,d^2) = W_1(\sum_{k=1}^{|T|}\bar{d}_k^1 \delta_{t_k}, \sum_{k=1}^{|T|}\bar{d}_k^2 \delta_{t_k} )
    \label{HOTT}
\end{equation}

Here, $\bar{d}^i$ represents document distributions over topics and the Dirac delta $\delta_{t_k}$ is a probability distribution supported on the corresponding topic $t_k$ and $W_1(d^1,d^2) = WMD(d^1,d^2)$  (\emph{WMD} being the Word Movers Distance). By truncating topics, the authors were able to reduce the computational time and make HOTT a competitive model against common methods. Their experiments show that although there is no uniformly best method, HOTT has on average the smallest error with respect to nBOW (normalized bag of words). More importantly, what was shown was that the process of truncating topics to improve computational time does not hinder the goal of obtaining high-quality distances. Interested readers will find in the paper more detailed reports about the setup and results of the experiments run.

\subsection{Domain Adaptation}
In this section we will cover \emph{Optimal Transport for Domain Adaptation} \cite{flamary2016optimal}.
In their paper, Flamary, \emph{et al.}, propose a regularized unsupervised optimal transportation model to perform an alignment of the representations in the source and target domains. By learning a transportation plan that matches the source and target PDFs, they constrained labeled samples of the same class during the transport. This helps solve the discrepancies (known as drift) in data distributions. 
\\
In real world problems, the drift that occurs between the source and target domains generally implies a change in marginal and conditional distributions. In this paper, the authors assume the domain drift is due to “an unknown, possibly nonlinear transformation of the input space $T: \Omega_s \rightarrow \Omega_t$ (omega is a measurable space, s is source, t is target). Because searching for T is an intractable problem and requires restrictions to become approximated. Here, the authors consider the problem of finding T the same as choosing a T such that one minimizes the transportation cost C(T):

\begin{equation}
    C(T) = \int_{\Omega_s} c(x, T(x))d\mu(x)
\end{equation}
where $c: \Omega_s \mathrm{ x } \Omega_t \rightarrow \mathbb{R}^+$ and $\mu(x)$ is a probability mass (or measure from x to T(x).)
\\
This is precisely the optimal transport problem. Then, to further improve the computational aspect of the model, a regularization component that preserves label information and sample neighborhood during the transportation is introduced. Now, the problem is as follows:
\begin{equation}
    \min\limits_{\pi \in \Pi} \langle\pi,C\rangle_F + \lambda\Omega_s(\pi)+\eta\Omega_c(\pi)
\end{equation}
where $\lambda \in \mathbb{R}$, $\eta \geq 0$, $\Omega_c(\cdot)$ is a class-based regularization term, and 
\begin{equation*}
    \Omega_s(\pi) = \sum_{i,j}\mathrm{\pi(i,j)log(i,j)}
\end{equation*}

This problem is solved using Algorithm \ref{alg:GenCondGradient}:

\begin{algorithm}
\DontPrintSemicolon
\caption{\textbf{Generalized Conditional Gradient}}
\label{alg:GenCondGradient}
Initialize: $k=0$, and $\pi^0 \in P$\;
repeat \;
\quad With $G \in \nabla f(\pi^k)$, solve $\pi^* = \underset{\pi\in B}{\mathrm{argmin}} \langle\pi,G\rangle_F + g(\pi)$ \;
\quad Find the optimal step $\alpha^k$, $\alpha^k = \underset{0\leq\alpha\leq 1}{\mathrm{argmin}} f(\pi^k + \alpha\Delta\pi)+g(\pi^k+\alpha\Delta\pi)$, with $\Delta\pi = \pi^* - \pi^k$ \;
\quad $\pi^{k+1} \leftarrow \pi^k + \alpha^k\Delta\pi$, set $k \leftarrow k+1$\;
until Convergence \;
\end{algorithm}
In the algorithm above, $f(\pi) = \langle\pi,C\rangle_F + \eta\Omega_c(\pi)$ and $g(\pi)=\lambda\Omega_s(\pi)$.
Using the assumption that $\Omega_c$ is differentiable, step 3 of the algorithm becomes
\begin{equation*}
    \pi^* = \underset{\pi\in Pi}{\mathrm{argmin}} \langle\pi, C+\eta\nabla\Omega_c(\pi^k)\rangle_F+\lambda\Omega_s(\pi)
\end{equation*}
By using a constrained optimal transport method, the overall performance was better than other state-of-the-art methods. Readers can find detailed reports on Table 1 in \cite{flamary2016optimal}.

For readers interested in domain adaptation, a varying approach to study heterogeneous domain adaptation problems using OT can be found in \cite{yan2018semi}.

\section{Future Research}
Further research will allow OT to be implemented in more areas and become more widely acceptable. The main problem with optimal transport is scaling onto higher dimensions. The optimal mappings that need to be solved are currently intractable in high-dimensions, which is where most of the current problems today lie. For example, Google's NLP model has roughly 1 trillion parameters. This type of problem is currently outside the scope of OT. Another interesting research topic is the use of optimal transport in approximating intractable distributions. This would compete with current known methods like KL-divergence and open up interesting opportunities when working with variational inference and/or expectation propagation. Another fundamental area to explore lies with the choice of using the Wasserstein distance. As shown throughout the paper, it is the most commonly used metric, but as one can see in Appendix 1, there are various others metrics, or distances, that may be used to replace W-distance. Interested readers can read more about them in Villani's Book, \emph{Optimal transport: old and new} \cite{villani2008optimal}. For further research from an applied perspective, one possibility is the use of the GWL framework explained in section \ref{Graphs} to improve on recommendation systems. On the other hand, all of the papers we have referenced above are quite novel in their applications and thus they all provide space for continuation or extension into more specific sub-fields within their respective context.

\section{Concluding Remarks} 
Throughout this survey, we have shown that Optimal Transport is seeing growing attention within the machine learning community due to its applicability in different areas. Although OT is becoming widely accepted in the machine learning world, it is deeply rooted in mathematics and so we extracted the most important topics so that interested readers can access only what is needed to have a high-level understanding of what is happening. These excerpts explain Kantorovich duality, entropic regularization, KL divergence, and Wasserstein barycenters. Although the applications of OT span a wide range, it is limited by computational challenges. Within this section we explored how using an entropic regularization term allowed for the formation of an algorithm that made OT problems computationally feasible and thus applicable. This takes us to the last section of this survey, the applications of optimal transport in machine learning. We began with computer vision, as it was one of the first applications of OT in ML. First, OT has been used to improve GANs by providing better statistical stability in low-dimensional data. Furthermore, since OT is one of the few methods that deal with the mass-splitting phenomenon, it allowed for many-to-one matching in pixel assignments which yielded a new approach to semantic correspondence with a 26\% performance improvement over state-of-the-art methods. The last application we covered with respect to computer vision was the use of W-distance to create a novel method for image interpolation called Convolutional Wasserstein Distance. Next, with respect to graphs, OT has allowed for the creation of the Gromov-Wasserstein Learning (GWL) algorithm which have also been shown to improve GANs. Other interesting areas that OT has shown promising results include neural architecture search, document representation, and domain adaptation. All of the papers we have analyzed and summarized will show that in some form (computational/accuracy) the use of OT has yielded better results than traditional methods. Although the computational inefficiencies are prevalent, the future for optimal transport in machine learning looks promising as more researchers become aware of this new intersection of areas.

\newpage
\section*{Appendix A}
The implementation of the conclusions of OT in machine learning rely mostly on the implementation of the various metrics that can be used as error measures in model tuning. The most notable ones arise from the reformulation or approximation of metrics into convex functionals that can be optimized by drawing on the many beautiful conclusions of convex geometry. Here, we recall a metric, many times called a distance, as a function $d(\cdot, \cdot): X \times X \to \mathbf{R}_+$, where $X$ is a metrizable space, that satisfies 
\begin{itemize}
    \item Positive Definite: $d(x, y) \geq 0$ with $d(x, y) = 0$ if, and only if, $x = y$
    \item Symmetric: $d(x,y) = d(y,x)$ for all $x,y \in X$
    \item Subadditive: $d(x,y) \leq d(x,z) + d(z,y)$ for all $x,y,z \in X$
\end{itemize}
Here, we want to note some different error estimates that come up in the OT literature as well as some that are traditionally used to compare probability distributions. The most notable comparison of probability measures in the OT literature is the p-Wasserstein Distance 
\begin{equation} \label{wpmetric}
    W^d_p (\mu_0 , \mu_1 ) =  \inf_{\pi \in \Pi(\mu_0 , \mu_1)} \bigg( \int_{M \times M} d(x,y)^p d\pi (x,y) \bigg)^{1/p}
\end{equation} 
In \ref{wpmetric}, d is a metric. We see from definition that it should very much act like a minimal $L^p$ distance on the space of probability measures. The most relevant choice of parameter p is $p=1,2$. This distance was formulated in the most general sense possible and it has a natural discrete formulation for discrete measures. Therefore, it allows for different contexts. For example, we saw the analog in the context of graphs as the Gromov-Wasserstein distance as: 

Let $(X,d_X,\mu_{X})$ and $(Y,d_Y,\mu_{Y})$ be two metric measure spaces, where $(X,d_X)$ is a compact metric space and $\mu_X$ is a probability measure on X (with $(Y,d_Y,\mu_{Y})$ defined in the same way). The Gromov-Wasserstein distance $d_{GW}(\mu_X,\mu_Y)$ is defined as
\\
\begin{equation}
\inf\limits_{\pi \in \Pi (\mu_X, \mu_Y)}\int\limits_{X \times Y} \int\limits_{X \times Y} L(x,y,x',y')d\pi(x,y)d\pi(x',y'),
\end{equation}

where $L(x,y,x',y') = |d_X(x,x') - d_Y(y,y')|$. Here, we see that the formulas look naturally similar. The Gromov-Wasserstein distance would be a particular choice of the 1-Wasserstein distance to a general metric space which can then be relaxed to be able to work with graphs as we saw before. 

The novelty in using OT in applications is principally the different error estimates. We recall some of the well-known distances that are traditionally used to compare probability measures: 

\begin{itemize}
    \item KL Divergence: \begin{align*}
        &KL(\pi|\kappa) \coloneqq \\ &\int\int_{M \times M} \pi(x,y) \bigg[\ln \frac{\pi(x,y)}{\kappa(x,y)} -1 \bigg] dxdy
    \end{align*}
    \item Hellinger distance: $H^2(\mu_0,\mu_1) = \frac{1}{2} \int (\sqrt{\frac{d\mu_0}{d\lambda}} - \sqrt{\frac{d\mu_1}{d\lambda}} )^2 d\lambda$, where $\mu_0,\mu_1$ are absolutely continuous with respect to $\lambda$ and $\frac{d\mu_0}{d\lambda}, \frac{d\mu_1}{d\lambda}$ denote the Radon-Nykodym derivatives, respectively. 
    \item Lèvy-Prokhorov distance:  $d_P(\mu_0,\mu_1) = \inf \{ \epsilon>0 ; \exists X,Y ; \inf \mathbb{P}[d(X,Y) >\epsilon] \leq \epsilon \}$ 
    \item Bounded Lipschitz distance (or Fortet-Mourier distance): $d_bL(\mu_0, \mu_1) = \sup \{ \int \phi d\mu_0 - \int \phi d\mu_1; ||\phi ||_\infty + || \phi ||_{\text{Lip}} \leq 1 \} $
    \item (in the case of nodes) Euclidean distance: $d(x,y) = \sqrt{(x-y)^2}$
\end{itemize}
We note that the Lèvy-Prokhorov and bounded Lipschitz distances can work in much the same way that the Wasserstein distance does. At the present, the Wasserstein distance proves useful because of it's capabilities in dealing with large distances and its convenient formulation in many problems such as the ones presented in this paper as well as others coming from partial differential equations. It's definition using infimum makes it easy to majorate. Its duality properties are useful--particularly in the case when $p=1$ as we see with the Kantorovich-Rubinstein distance where it is defined as an equivalence to its dual:
\begin{equation}
    W_1(\mu_0, \mu_1) = \sup_{||\phi||_{\text{Lip}} \leq 1} \bigg\{ \int_X \phi d\mu_0 - \int_x \phi d\mu_1 \bigg\}
\end{equation}
The interested reader can read more about the different distances in \cite{rachev1991probability,villani2008optimal}

As we presently see in this paper, we notice that much of the work on the optimal transport in machine learning is in the reformulation of the algorithms, which classically used the traditional distances, into new versions that use the Wasserstein distance. Then, a lot of the work is done in dealing with the computational inefficiency of the Wasserstein distance. Moving forward, the authors think that many machine learning algorithms will implement some of the "deeper" features of the optimal transport theory to improve such algorithms after their best formulation becomes abundantly clear. 

\section*{Appendix B}
For the readers interested in papers that apply OT in machine learning, here are a few more references to be considered. First we have OT in GANS:
\begin{itemize}
    \item \textit{A geometric view of optimal transportation and generative model} \cite{lei2019geometric}
\end{itemize}
Next, for semantic correspondence and NLP we have:
\begin{itemize}
    \item \textit{Improving sequence-to-sequence learning via optimal transport} \cite{chen2019improving}
\end{itemize}
Lastly, on domain adaptation we have:
\begin{itemize}
    \item \textit{Joint distribution optimal transportation for domain adaptation} \cite{courty2017joint}
    \item \textit{Theoretical analysis of domain adaptation with optimal transport} \cite{redko2017theoretical}
\end{itemize}

\section*{Acknowledgement}
This material is based upon Luiz Manella Pereira's work supported by the U.S. Department of Homeland Security under Grant Award Number, 2017‐ST‐062‐000002. The views and conclusions contained in this document are those of the authors and should not be interpreted as necessarily representing the official policies, either expressed or implied, of the U.S. Department of Homeland Security.

\printbibliography

@article{chen2019improving,
  title={Improving sequence-to-sequence learning via optimal transport},
  author={Chen, Liqun and Zhang, Yizhe and Zhang, Ruiyi and Tao, Chenyang and Gan, Zhe and Zhang, Haichao and Li, Bai and Shen, Dinghan and Chen, Changyou and Carin, Lawrence},
  journal={arXiv preprint arXiv:1901.06283},
  year={2019}
}

@article{courty2017joint,
  title={Joint distribution optimal transportation for domain adaptation},
  author={Courty, Nicolas and Flamary, R{\'e}mi and Habrard, Amaury and Rakotomamonjy, Alain},
  journal={arXiv preprint arXiv:1705.08848},
  year={2017}
}

@inproceedings{redko2017theoretical,
  title={Theoretical analysis of domain adaptation with optimal transport},
  author={Redko, Ievgen and Habrard, Amaury and Sebban, Marc},
  booktitle={Joint European Conference on Machine Learning and Knowledge Discovery in Databases},
  pages={737--753},
  year={2017},
  organization={Springer}
}

@article{salimans2018improving,
  title={Improving GANs using optimal transport},
  author={Salimans, Tim and Zhang, Han and Radford, Alec and Metaxas, Dimitris},
  journal={arXiv preprint arXiv:1803.05573},
  year={2018}
}

@article{peyre2019computational,
  title={Computational optimal transport},
  author={Peyr{\'e}, Gabriel and Cuturi, Marco and others},
  journal={Foundations and Trends{\textregistered} in Machine Learning},
  volume={11},
  number={5-6},
  pages={355--607},
  year={2019},
  publisher={Now Publishers, Inc.}
}

@article{lei2019geometric,
  title={A geometric view of optimal transportation and generative model},
  author={Lei, Na and Su, Kehua and Cui, Li and Yau, Shing-Tung and Gu, Xianfeng David},
  journal={Computer Aided Geometric Design},
  volume={68},
  pages={1--21},
  year={2019},
  publisher={Elsevier}
}

@article{monge1781memoire,
  title={M{\'e}moire sur la th{\'e}orie des d{\'e}blais et des remblais},
  author={Monge, Gaspard},
  journal={Histoire de l'Acad{\'e}mie Royale des Sciences de Paris},
  year={1781}
}

@article{mather1989minimal,
  title={Minimal measures},
  author={Mather, John N},
  journal={Commentarii Mathematici Helvetici},
  volume={64},
  number={1},
  pages={375--394},
  year={1989},
  publisher={Springer}
}

@article{bernard2004optimal,
  title={Optimal mass transportation and Mather theory},
  author={Bernard, Patrick and Buffoni, Boris},
  journal={arXiv preprint math/0412299},
  year={2004}
}

@article{peleg1989unified,
  title={A unified approach to the change of resolution: Space and gray-level},
  author={Peleg, Shmuel and Werman, Michael and Rom, Hillel},
  journal={IEEE Transactions on Pattern Analysis and Machine Intelligence},
  volume={11},
  number={7},
  pages={739--742},
  year={1989},
  publisher={IEEE}
}

@inproceedings{rubner1998metric,
  title={A metric for distributions with applications to image databases},
  author={Rubner, Yossi and Tomasi, Carlo and Guibas, Leonidas J},
  booktitle={Sixth International Conference on Computer Vision (IEEE Cat. No. 98CH36271)},
  pages={59--66},
  year={1998},
  organization={IEEE}
}

@article{rubner2000earth,
  title={The earth mover's distance as a metric for image retrieval},
  author={Rubner, Yossi and Tomasi, Carlo and Guibas, Leonidas J},
  journal={International journal of computer vision},
  volume={40},
  number={2},
  pages={99--121},
  year={2000},
  publisher={Springer}
}

@article{haker2004optimal,
  title={Optimal mass transport for registration and warping},
  author={Haker, Steven and Zhu, Lei and Tannenbaum, Allen and Angenent, Sigurd},
  journal={International Journal of computer vision},
  volume={60},
  number={3},
  pages={225--240},
  year={2004},
  publisher={Springer}
}

@article{haker2003monge,
  title={On the Monge-Kantorovich problem and image warping},
  author={Haker, Steven and Tannenbaum, Allen},
  journal={IMA Volumes in Mathematics and its Applications},
  volume={133},
  pages={65--86},
  year={2003},
  publisher={New York; Springer; 1999}
}

@article{gangbo2000shape,
  title={Shape recognition via Wasserstein distance},
  author={Gangbo, Wilfrid and McCann, Robert J},
  journal={Quarterly of Applied Mathematics},
  pages={705--737},
  year={2000},
  publisher={JSTOR}
}

@article{ahmad2003geometry,
  title={The geometry of shape recognition via the Monge-Kantorovich optimal transport problem.},
  author={Ahmad, Najma},
  year={2003}
}

@article{sandler2011nonnegative,
  title={Nonnegative matrix factorization with earth mover's distance metric for image analysis},
  author={Sandler, Roman and Lindenbaum, Michael},
  journal={IEEE Transactions on Pattern Analysis and Machine Intelligence},
  volume={33},
  number={8},
  pages={1590--1602},
  year={2011},
  publisher={IEEE}
}

@inproceedings{solomon2014wasserstein,
  title={Wasserstein propagation for semi-supervised learning},
  author={Solomon, Justin and Rustamov, Raif and Guibas, Leonidas and Butscher, Adrian},
  booktitle={International Conference on Machine Learning},
  pages={306--314},
  year={2014}
}

@inproceedings{ambrogioni2018wasserstein,
  title={Wasserstein variational inference},
  author={Ambrogioni, Luca and G{\"u}{\c{c}}l{\"u}, Umut and G{\"u}{\c{c}}l{\"u}t{\"u}rk, Ya{\u{g}}mur and Hinne, Max and van Gerven, Marcel AJ and Maris, Eric},
  booktitle={Advances in Neural Information Processing Systems},
  pages={2473--2482},
  year={2018}
}

@inproceedings{cuturi2013sinkhorn,
  title={Sinkhorn distances: Lightspeed computation of optimal transport},
  author={Cuturi, Marco},
  booktitle={Advances in neural information processing systems},
  pages={2292--2300},
  year={2013}
}

@article{santambrogiooptimal,
  title={Optimal Transport meets Probability, Statistics and Machine Learning},
  author={Santambrogio, Filippo}
}

@article{solomon2015convolutional,
  title={Convolutional wasserstein distances: Efficient optimal transportation on geometric domains},
  author={Solomon, Justin and De Goes, Fernando and Peyr{\'e}, Gabriel and Cuturi, Marco and Butscher, Adrian and Nguyen, Andy and Du, Tao and Guibas, Leonidas},
  journal={ACM Transactions on Graphics (TOG)},
  volume={34},
  number={4},
  pages={1--11},
  year={2015},
  publisher={ACM New York, NY, USA}
}

@article{varadhan1967behavior,
  title={On the behavior of the fundamental solution of the heat equation with variable coefficients},
  author={Varadhan, Sathamangalam R Srinivasa},
  journal={Communications on Pure and Applied Mathematics},
  volume={20},
  number={2},
  pages={431--455},
  year={1967},
  publisher={Wiley Online Library}
}

@article{arjovsky2017wasserstein,
  title={Wasserstein gan},
  author={Arjovsky, Martin and Chintala, Soumith and Bottou, L{\'e}on},
  journal={arXiv preprint arXiv:1701.07875},
  year={2017}
}

@article{BellemareDDMLHM17,
  author    = {Marc G. Bellemare and
               Ivo Danihelka and
               Will Dabney and
               Shakir Mohamed and
               Balaji Lakshminarayanan and
               Stephan Hoyer and
               R{\'{e}}mi Munos},
  title     = {The Cramer Distance as a Solution to Biased Wasserstein Gradients},
  journal   = {CoRR},
  volume    = {abs/1705.10743},
  year      = {2017},
  url       = {http://arxiv.org/abs/1705.10743},
  archivePrefix = {arXiv},
  eprint    = {1705.10743},
  bibsource = {dblp computer science bibliography, https://dblp.org}
}

@misc{genevay2017learning,
      title={Learning Generative Models with Sinkhorn Divergences}, 
      author={Aude Genevay and Gabriel Peyré and Marco Cuturi},
      year={2017},
      eprint={1706.00292},
      archivePrefix={arXiv},
      primaryClass={stat.ML}
}

@inproceedings{liu2020semantic,
  title={Semantic Correspondence as an Optimal Transport Problem},
  author={Liu, Yanbin and Zhu, Linchao and Yamada, Makoto and Yang, Yi},
  booktitle={Proceedings of the IEEE/CVF Conference on Computer Vision and Pattern Recognition},
  pages={4463--4472},
  year={2020}
}

@article{agueh2011barycenters,
  title={Barycenters in the Wasserstein space},
  author={Agueh, Martial and Carlier, Guillaume},
  journal={SIAM Journal on Mathematical Analysis},
  volume={43},
  number={2},
  pages={904--924},
  year={2011},
  publisher={SIAM}
}

@article{cuturi2014fast,
  title={Fast computation of Wasserstein barycenters},
  author={Cuturi, Marco and Doucet, Arnaud},
  year={2014},
  publisher={Journal of Machine Learning Research}
}

@article{xu2019gromov,
  title={Gromov-wasserstein learning for graph matching and node embedding},
  author={Xu, Hongteng and Luo, Dixin and Zha, Hongyuan and Carin, Lawrence},
  journal={arXiv preprint arXiv:1901.06003},
  year={2019}
}

@inproceedings{yurochkin2019hierarchical,
  title={Hierarchical optimal transport for document representation},
  author={Yurochkin, Mikhail and Claici, Sebastian and Chien, Edward and Mirzazadeh, Farzaneh and Solomon, Justin M},
  booktitle={Advances in Neural Information Processing Systems},
  pages={1601--1611},
  year={2019}
}

@article{flamary2016optimal,
  title={Optimal transport for domain adaptation},
  author={Flamary, R and Courty, N and Tuia, D and Rakotomamonjy, A},
  journal={IEEE Trans. Pattern Anal. Mach. Intell},
  year={2016}
}

@article{kolouri2020wasserstein,
  title={Wasserstein Embedding for Graph Learning},
  author={Kolouri, Soheil and Naderializadeh, Navid and Rohde, Gustavo K and Hoffmann, Heiko},
  journal={arXiv preprint arXiv:2006.09430},
  year={2020}
}

@inproceedings{kandasamy2018neural,
  title={Neural architecture search with bayesian optimisation and optimal transport},
  author={Kandasamy, Kirthevasan and Neiswanger, Willie and Schneider, Jeff and Poczos, Barnabas and Xing, Eric P},
  booktitle={Advances in neural information processing systems},
  pages={2016--2025},
  year={2018}
}

@book{villani2003topics,
  title={Topics in optimal transportation},
  author={Villani, C{\'e}dric},
  number={58},
  year={2003},
  publisher={American Mathematical Soc.}
}

@inproceedings{yan2018semi,
  title={Semi-Supervised Optimal Transport for Heterogeneous Domain Adaptation.},
  author={Yan, Yuguang and Li, Wen and Wu, Hanrui and Min, Huaqing and Tan, Mingkui and Wu, Qingyao},
  booktitle={IJCAI},
  pages={2969--2975},
  year={2018}
}

@article{scetbon2020handling,
  title={Handling multiple costs in optimal transport: Strong duality and efficient computation},
  author={Scetbon, Meyer and Meunier, Laurent and Atif, Jamal and Cuturi, Marco},
  journal={arXiv preprint arXiv:2006.07260},
  year={2020}
}

@article{bunne2019learning,
  title={Learning generative models across incomparable spaces},
  author={Bunne, Charlotte and Alvarez-Melis, David and Krause, Andreas and Jegelka, Stefanie},
  journal={arXiv preprint arXiv:1905.05461},
  year={2019}
}

@book{villani2008optimal,
  title={Optimal transport: old and new},
  author={Villani, C{\'e}dric},
  volume={338},
  year={2008},
  publisher={Springer Science \& Business Media}
}

@article{santambrogio2015optimal,
  title={Optimal transport for applied mathematicians},
  author={Santambrogio, Filippo},
  journal={Birk{\"a}user, NY},
  volume={55},
  number={58-63},
  pages={94},
  year={2015},
  publisher={Springer}
}

@article{solomon2014earth,
  title={Earth mover's distances on discrete surfaces},
  author={Solomon, Justin and Rustamov, Raif and Guibas, Leonidas and Butscher, Adrian},
  journal={ACM Transactions on Graphics (TOG)},
  volume={33},
  number={4},
  pages={1--12},
  year={2014},
  publisher={ACM New York, NY, USA}
}

@article{merigot2020optimal,
  title={Optimal transport: discretization and algorithms},
  author={Merigot, Quentin and Thibert, Boris},
  journal={arXiv preprint arXiv:2003.00855},
  year={2020}
}

@article{vialard2019elementary,
  title={An elementary introduction to entropic regularization and proximal methods for numerical optimal transport},
  author={Vialard, Fran{\c{c}}ois-Xavier},
  year={2019}
}

@inproceedings{genevay2016stochastic,
  title={Stochastic optimization for large-scale optimal transport},
  author={Genevay, Aude and Cuturi, Marco and Peyr{\'e}, Gabriel and Bach, Francis},
  booktitle={Advances in neural information processing systems},
  pages={3440--3448},
  year={2016}
}

@book{rachev1991probability,
title={Probability Metrics and the Stability of Stochastic Models}, author={Rachev, S.T.},
isbn={9780471928775},
lccn={90043733},
series={Wiley Series in Probability and Statistics - Applied Probability and
Statistics Section}, url={https://books.google.com/books?id=5grvAAAAMAAJ}, year={1991},
publisher={Wiley}
}

\end{document}